\documentclass[10pt]{article}

\usepackage[margin=0.75in]{geometry}
\usepackage{times}
\usepackage{microtype}
\usepackage{amsmath,amssymb,amsthm}
\usepackage{mathtools}
\usepackage{enumitem}
\usepackage{csquotes}
\usepackage[hidelinks]{hyperref}

\usepackage[
  backend=biber,
  style=authoryear,
  natbib=true,
  maxcitenames=2,
  maxbibnames=99
]{biblatex}

\theoremstyle{definition}
\newtheorem{definition}{Definition}
\newtheorem{assumption}{Assumption}
\theoremstyle{plain}
\newtheorem{theorem}{Theorem}
\newtheorem{lemma}{Lemma}
\newtheorem{proposition}{Proposition}
\newtheorem{corollary}{Corollary}
\theoremstyle{remark}
\newtheorem{remark}{Remark}

\newcommand{\Pow}{\mathcal{P}}
\newcommand{\Nat}{\mathbb{N}}

\newcommand{\Truth}[1]{\mathrm{T}(#1)}      
\newcommand{\Ext}[1]{\mathrm{Ext}(#1)}      
\newcommand{\Lang}[1]{\mathcal{L}_{#1}}     
\newcommand{\Enc}{\mathrm{Enc}}             
\newcommand{\WinMap}{\mathrm{W}}            
\newcommand{\Occur}{\mathrm{Occur}}         
\newcommand{\CoInst}{\mathrm{CoInst}}       
\newcommand{\card}[1]{\lvert #1\rvert}

\newcommand{\Intuition}{\paragraph{Intuition.}}
\newcommand{\ProofIntuition}{\paragraph{Proof intuition.}}

\newcommand{\DiamondOp}{\Diamond}
\newcommand{\BoxOp}{\Box}

\newcommand{\Ground}[2]{\mathrm{Ground}_{#1\leftarrow #2}}

\title{\vspace{-0.5em}A Mind Cannot Be Smeared Across Time\vspace{-0.4em}}
\author{Michael Timothy Bennett }
\date{2026.1.11}

\begin{document}
\maketitle

\begin{center}
\begin{minipage}{0.7\textwidth}
\begin{abstract}
Whether machines can be conscious depends not only on what they compute, but \emph{when} they compute it. Most deployed artificial systems realise their functions via sequential or time-multiplexed updates, yet a moment of conscious experience feels unified and simultaneous. I prove that this difference matters. I augment Stack Theory with algebraic laws relating within time-window constraint satisfaction to conjunction. I introduce a temporal semantics over windowed trajectories $\tau_{\Delta}$ and prove that existential temporal realisation $\Diamond_{\Delta}$ does not preserve conjunction. A system can realise all the ingredients of experience across time without ever instantiating the experienced conjunction itself. I then distinguish two postulates, Chord and Arpeggio. Chord is the position that conscious unity requires \textit{objective co-instantiation} of the grounded conjunction within the window, like a musical chord. Arpeggio only needs the ingredients to \textit{occur} within window, like a melody. I formalise concurrency-capacity to measure what is needed to satisfy co-instantiation. Finally, I review neurophysiological evidence suggesting that consciousness depends on phase synchrony and effective connectivity, and that loss of consciousness is associated with its breakdown. Under Chord, software consciousness on strictly sequential substrates is impossible for contents whose grounding requires two or more simultaneous contributors. The hardware matters.
\end{abstract}
\end{minipage}
\end{center}

\section{Introduction}

A central open question in machine consciousness is whether \emph{behaviourally} and \emph{functionally} matched systems must be matched in \emph{phenomenal realisation}. Stack Theory approaches this by formalising cognitive processes in terms of abstraction layers \citep{bennett2025thesis,bennett2023c,bennettwelshciaunica2024}. Various aspects of incumbent theories of consciousness are formalised \citep{seth2022,block1995,chalmers1995,rosenthal1986two,solms2021,baars1988}, organised into a hierarchy of causal identities \citep{bennett2023c}, and derived as a necessary consequence of a generalisation-optimal learning theory that makes valence primary \citep{bennett2025e,bennett2025d,bennett2023b,bennett2022c}. Those necessary ingredients of consciousness are beyond the scope of this paper. Here I am concerned with a question of sufficiency. Stack Theory shows how the behaviour of two systems can \textit{appear} identical at one level of abstraction, while being different in the levels below. Conscious experience can be purely functional, but still depend on more than surface level function---not just on \textit{what} is computed, but \textit{when} and \textit{how}. The ingredients of what looks like a single experienced moment at a high level of abstraction may be smeared across objective micro-time at a lower level, such that no objective time-slice contains the full grounded conjunction. If the ingredients of a subjective moment need not synchronise in objective time, then current hardware suffices to host machine consciousness---and a ``liquid-brained'' swarm intelligence \citep{sole2019} like an ant colony can be conscious. But if synchronised broadcast and causal co-dependence matter, then far fewer systems qualify. This question of architectural sufficiency is the \emph{Temporal Gap}.

The Temporal Gap was proposed in earlier work \citep{bennett2025thesis}, but was not rigorously formalised. Here I add a \emph{Stack-Time Semantics} module to Stack Theory, including induced layer time, window environments, temporal lifting operators, and algebraic laws relating ``within-window'' satisfaction to conjunction.
Using this, I refine the Temporal Gap into a testable architectural question about synchrony and concurrency.

\paragraph{Relation to existing consciousness theories.}
Beyond Stack Theory, many scientific theories emphasise some form of \emph{integration} or \emph{unity} across time and causal organisation: global workspace theory \citep{baars1988,dehaene2001}, integrated information \citep{tononi2004,oizumi2014}, higher-order representation \citep{rosenthal1986two}, attention-schema modelling \citep{graziano2015ast}, predictive-processing and active-inference \citep{friston2010}, and more general analyses of time and consciousness \citep{James1890,varela1999specious}. 
The aim here is not to adjudicate among theories, but to extract a common formal issue: if phenomenal content is modelled as a \emph{conjunction} of grounded ingredients, what does it mean for that conjunction to be realised \emph{within the temporal extent of a subjective moment}?

\subsection{Results}
\begin{enumerate}[leftmargin=*, itemsep=0.35em]
\item \textbf{Compositional grounding.} Any higher-layer statement can be grounded compositionally to the base layer preserving truth conditions (Theorem~\ref{thm:comp_ground}).
\item \textbf{Subjective vs objective time.} Higher-layer ticks are maximal objective-time blocks of constant encoding, and can be arbitrarily sparse (Proposition~\ref{prop:sparse_ticks}).
\item \textbf{Temporal lifting algebra.} The core non-commutation result underpinning the Temporal Gap (Theorem~\ref{thm:diamond}).
\item \textbf{Chord vs Arpeggio postulates.} Both postulates stated in terms of a higher-layer moment statement and its base-layer grounding (Definitions~\ref{def:tgstrong} and~\ref{def:tgweak}).
\item \textbf{Concurrency capacity threshold.} Architectures with limited simultaneous contribution can satisfy ingredient-wise occurrence while forbidding co-instantiation beyond capacity (Theorem~\ref{thm:capacity}).
\item \textbf{Evidence supporting Chord.} Co-instantiation requirements are connected to evidence linking consciousness to synchrony and effective connectivity (Section~\ref{sec:neuro}).
\end{enumerate}

\section{Stack Theory Primitives}
\label{sec:stack_prims}

I follow the core notation of Stack Theory \citep{bennett_appendix,bennett2025thesis}.

\begin{definition}[Environment and programs]
An \emph{environment} is a nonempty set $\Phi$ of mutually exclusive states.
A \emph{program} is a set $p\subseteq\Phi$.
A program $p$ is true at $\phi\in\Phi$ iff $\phi\in p$.
Let $P:=2^\Phi$ be the set of all programs.
A \emph{vocabulary} is any finite $\mathfrak{v}\subseteq P$.
\end{definition}

\Intuition An environment is the set of possible states. A program is a set of states, each state representing a point in time and a smallest possible change in an environment, equating time with change \citep{bennett2025thesis}. A program is true at a state when the state is in that set. Conjunction is set intersection.

\begin{definition}[Statements, truth sets, extensions]
The \emph{language} generated by $\mathfrak{v}$ is
\[
\Lang{\mathfrak{v}}:=\left\{\,l\subseteq\mathfrak{v}\;:\;l\ \text{is finite and }\bigcap_{p\in l}p\neq\emptyset\,\right\}.
\]
Elements $l\in\Lang{\mathfrak{v}}$ are (conjunctive) \emph{statements}.
The \emph{truth set} of $l$ is
\[
\Truth{l}:=\bigcap_{p\in l}p\subseteq \Phi.
\]
I say $l$ is true at $\phi$ iff $\phi\in\Truth{l}$.
By convention, $\Truth{\emptyset}=\Phi$.\\

\noindent A \emph{completion} of $l$ is any $l'\in\Lang{\mathfrak{v}}$ with $l\subseteq l'$.
Define the \emph{extension} of $l$ by $\Ext{l}:=\{\,l'\in\Lang{\mathfrak{v}}:l\subseteq l'\,\}$.
For $X\subseteq\Lang{\mathfrak{v}}$, write $\Ext{X}:=\bigcup_{x\in X}\Ext{x}$.
\end{definition}

\Intuition A statement is a finite bundle of programs that must all be true at once. A completion adds more programs, making the statement more specific. The extension collects all completions that still include the original statement.

\begin{definition}[Tasks and policies]
\label{def:tasks_policies}
A \emph{task} is a pair $\alpha=\langle I_\alpha,O_\alpha\rangle$ where
$ I_\alpha\subseteq\Lang{\mathfrak{v}}$ is a set of admissible input statements and $O_\alpha\subseteq\Ext{I_\alpha}$ is a set of correct output statements.
A \emph{policy} is any statement $\pi\in\Lang{\mathfrak{v}}$.
A policy is \emph{correct} for $\alpha$ iff
\[
\Ext{I_\alpha}\cap \Ext{\pi}=O_\alpha.
\]
Let $\Pi_\alpha$ denote the set of correct policies for $\alpha$.
\end{definition}

\Intuition A task specifies which inputs are allowed and which outputs count as correct. A policy is correct when it selects exactly the correct outputs for the allowed inputs.

\section{Grounding Stack Theory}
\label{sec:stack_grounding}

\begin{definition}[Abstractor]
\label{def:abstractor}
Fix a base vocabulary $\mathfrak{v}\subseteq P$ and a policy $\pi\in\Lang{\mathfrak{v}}$.
The induced \emph{abstractor} is the function
\[
\mathfrak{f}(\mathfrak{v},\pi)
:=\left\{\,q\in P\;:\;\exists o\in \Ext{\pi}\ \text{s.t.}\ \bigcap_{r\in o} r=q\,\right\}.
\]
Thus $\mathfrak{f}(\mathfrak{v},\pi)$ is a new vocabulary consisting of truth sets of completions compatible with $\pi$.
\end{definition}

\Intuition The abstractor takes completions of the policy and turns each into one set of states. The collection of these sets forms the vocabulary of the next layer---a compression that maps completions to equivalent programs.

\begin{definition}[Stack]
A \emph{stack of depth $m$} is a sequence of uninstantiated task functions
$\langle \lambda^0,\lambda^1,\dots,\lambda^m\rangle$
such that for each $i<m$, $\lambda^{i+1}$ is a restriction of $\lambda^i$ (written $\lambda^{i+1}\sqsubset \lambda^i$). Each $\lambda^i$ assigns to each vocabulary $\mathfrak{v}$ a task $\lambda^i(\mathfrak{v})$. The notation $\Pi_{\lambda^i(\mathfrak{v})}$ denotes the set of correct policies for that task.
A \emph{stack state} is a pair of sequences
\[
(\pi^0,\dots,\pi^{m-1})\quad\text{and}\quad (\mathfrak{v}^0,\dots,\mathfrak{v}^{m})
\]
satisfying, for each $i<m$,
\[
\mathfrak{v}^{i+1}=\mathfrak{f}(\mathfrak{v}^{i},\pi^{i})
\qquad\text{and}\qquad
\pi^{i}\in\Pi_{\lambda^i(\mathfrak{v}^i)}.
\]
\end{definition}

\Intuition At each layer you pick a policy that solves the task for that layer. That policy also defines the next vocabulary, the higher level description of what is going on.

\subsection{Grounding a Higher-Layer into Lower Layers}

\begin{lemma}[Grounding through one abstractor]
\label{lem:ground_one}
Let $\mathfrak{v}\subseteq P$ and $\pi\in\Lang{\mathfrak{v}}$, and set $\mathfrak{v}'=\mathfrak{f}(\mathfrak{v},\pi)$.
Then for every program $p\in\mathfrak{v}'$ there exists a completion $o_p\in\Ext{\pi}$ such that $\Truth{o_p}=p$.
Moreover, for any statement $l\in\Lang{\mathfrak{v}'}$ there exists a statement $\mathrm{ground}_\pi(l)\in\Lang{\mathfrak{v}}$ with
$\Truth{l}=\Truth{\mathrm{ground}_\pi(l)}$.
\end{lemma}

\noindent\textit{Proof.} See Supplementary Material. Each macro-ingredient has a lower-layer completion with the same truth set; replacing ingredients and intersecting preserves truth conditions. \qed

\subsection{Compositional Grounding}

\begin{definition}[Compositional grounding across a stack]
\label{def:comp_ground}
Fix a stack with vocabularies $\mathfrak{v}^0,\dots,\mathfrak{v}^m$ and policies $\pi^0,\dots,\pi^{m-1}$.
For a statement $l^m\in\Lang{\mathfrak{v}^m}$, define grounded statements $g^m,\dots,g^0$ recursively by
\begin{align*}
g^m &:= l^m,\\
g^i &:= \mathrm{ground}_{\pi^{i}}(g^{i+1})\in\Lang{\mathfrak{v}^{i}}
\quad (i=m{-}1,\dots,0).
\end{align*}
Define the \emph{compositional grounding} map $\Ground{0}{m}$ by $\Ground{0}{m}(l^m):=g^0$.
\end{definition}

\Intuition Grounding works like compiling a high level statement into lower level pieces. Replace each high level ingredient with a lower level statement that means the same thing. Repeating this down the stack gives a base level conjunction true in exactly the same cases.

\begin{theorem}[Compositional grounding preserves truth conditions]
\label{thm:comp_ground}
For any stack and any statement $l^m\in\Lang{\mathfrak{v}^m}$,
\[
\Truth{\Ground{0}{m}(l^m)}=\Truth{l^m}.
\]
\end{theorem}

\begin{proof}
Apply Lemma~\ref{lem:ground_one} iteratively:
$\Truth{g^{i}}=\Truth{g^{i+1}}$ for each $i<m$, hence $\Truth{g^0}=\Truth{g^m}=\Truth{l^m}$.
\end{proof}

\section{Stack-Time Semantics Module}
\label{sec:stack_time}

This section defines induced layer time, window semantics, and temporal lifting operators as a reusable addendum to Stack Theory.

\subsection{Objective Time and Layer Time as a Quotient}
\label{sec:layer_time}

\begin{definition}[Objective time and trajectories]
Objective time is represented by a discrete index $t\in\Nat$.
An \emph{objective trajectory} is a function $\tau:\Nat\to\Phi$ such that
$\tau(t{+}1)\neq \tau(t)$ for all $t\in\Nat$.
\end{definition}

\Intuition Objective time is a step counter for objective change.
If nothing changes then there is no next objective index.
Two events are simultaneous when they occur at the same objective index.

\begin{definition}[Encoding map]
For a vocabulary $\mathfrak{v}\subseteq P$ and a state $\phi\in\Phi$, define
$\Enc_{\mathfrak{v}}(\phi):=\{\,p\in\mathfrak{v}:\phi\in p\,\}\subseteq \mathfrak{v}$.
\end{definition}

\Intuition Given a state and a vocabulary, the encoding collects all programs in that vocabulary true at that state---the most detailed true statement in that vocabulary.

\begin{lemma}[Encoding yields the maximal true statement]
\label{lem:enc}
For any $\mathfrak{v}$ and $\phi\in\Phi$, $\Enc_{\mathfrak{v}}(\phi)\in\Lang{\mathfrak{v}}$ and $\phi\in\Truth{\Enc_{\mathfrak{v}}(\phi)}$.
Moreover, if $l\in\Lang{\mathfrak{v}}$ is true at $\phi$, then $l\subseteq \Enc_{\mathfrak{v}}(\phi)$.
\end{lemma}

\noindent\textit{Proof.} See Supplementary Material. \qed

\begin{definition}[Induced layer trajectory and layer time]
Given an objective trajectory $\tau$ and a vocabulary $\mathfrak{v}$, define the induced layer trajectory
$\tau^{(\mathfrak{v})}(t):=\Enc_{\mathfrak{v}}(\tau(t))\in\Lang{\mathfrak{v}}$.
Define the \emph{layer-tick map} $\kappa_{\mathfrak{v}}:\Nat\to\Nat$ by $\kappa_{\mathfrak{v}}(0)=0$ and, for $t>0$,
\[
\kappa_{\mathfrak{v}}(t):=
\begin{cases}
\kappa_{\mathfrak{v}}(t{-}1) &\text{if }\tau^{(\mathfrak{v})}(t)=\tau^{(\mathfrak{v})}(t{-}1),\\
\kappa_{\mathfrak{v}}(t{-}1)+1 &\text{otherwise.}
\end{cases}
\]
Define $t\approx_{\mathfrak{v}} t'$ iff $\kappa_{\mathfrak{v}}(t)=\kappa_{\mathfrak{v}}(t')$.
The \emph{layer time} is $\Nat/\!\approx_{\mathfrak{v}}$: each equivalence class is a maximal contiguous block of objective times on which $\tau^{(\mathfrak{v})}$ is constant.
\end{definition}

\Intuition For each objective time step, write down the encoding. If it stays the same for several steps, treat them as one higher level moment. Layer time advances only when something visible at that level changes.

\begin{proposition}[Layer time can be arbitrarily sparse]
\label{prop:sparse_ticks}
For any $N\in\Nat$ there exist $\Phi,\mathfrak{v},\tau$ such that the induced trajectory $\tau^{(\mathfrak{v})}$ changes at most once in the first $N$ objective steps.
\end{proposition}

\noindent\textit{Proof.} See Supplementary Material. Choose a vocabulary that cannot distinguish many distinct objective microstates; the macrostate stays constant. \qed

\subsection{Time and Lifting Operators}
\label{sec:window}

A subjective moment is commonly modelled as having nonzero temporal extent (the ``specious present'') \citep{James1890,varela1999specious}.
I introduce temporal operators to formalise satisfaction \emph{within} such a window.

\begin{definition}[Ingredient-wise vs co-instantiated satisfaction]
\label{def:window_sat}
Fix a windowing map $\WinMap$, an objective trajectory $\tau$, and a window index $t$.
For a program $p\subseteq\Phi$, define
\[
\Occur_{\WinMap}(p,\tau,t)\iff \exists u\in\WinMap(t)\ \big(\tau(u)\in p\big).
\]
For a statement $l\in\Lang{\mathfrak{v}}$, define
\[
\Occur_{\WinMap}(l,\tau,t)\iff \forall p\in l\ \Occur_{\WinMap}(p,\tau,t).
\]
Define co-instantiation of the conjunction by
\[
\CoInst_{\WinMap}(l,\tau,t)\iff \exists u\in\WinMap(t)\ \big(\tau(u)\in \Truth{l}\big).
\]
\end{definition}

\Intuition Ingredient-wise means every ingredient becomes true at least once somewhere in the window. Co-instantiated means there is a time step in the window where all ingredients are true together.

\begin{remark}[Quantifier structure]
For conjunctive $l$, ingredient-wise satisfaction is $\forall p\in l\ \exists u\in\WinMap(t)$, while co-instantiation is $\exists u\in\WinMap(t)\ \forall p\in l$.
The Temporal Gap amounts to whether conscious unity requires ingredient-wise satisfaction or co-instantiation.
\end{remark}

\begin{definition}[Window environments and window trajectories]
Fix $\Delta\in\Nat$.
Define the \emph{window environment} $\Phi^\Delta:=\Phi^{\{0,1,\dots,\Delta\}}\cong \Phi^{\Delta+1}$ whose elements are windows $\sigma=(\phi_0,\dots,\phi_\Delta)$.
Given an objective trajectory $\tau$, define the induced window trajectory
$\tau_{\Delta}(t):=\big(\tau(t),\tau(t{+}1),\dots,\tau(t{+}\Delta)\big)\in\Phi^\Delta$.
\end{definition}

\Intuition Treat a whole window of $\Delta{+}1$ consecutive states as a single new state.

\begin{definition}[Temporal lifts of programs]
For a program $p\subseteq\Phi$, define the lifted programs on $\Phi^\Delta$:
\[
\begin{aligned}
\DiamondOp_\Delta p &:= \{\,\sigma\in\Phi^\Delta : \exists k\le \Delta\ \text{s.t.}\ \sigma(k)\in p\,\},\\
\BoxOp_\Delta p &:= \{\,\sigma\in\Phi^\Delta : \forall k\le \Delta,\ \sigma(k)\in p\,\}.
\end{aligned}
\]
\end{definition}

\Intuition $\Diamond_\Delta p$ means $p$ is true at least once in the window. $\Box_\Delta p$ means $p$ is true at every step in the window.

\begin{definition}[Lifted statements]
For a statement $l\in\Lang{\mathfrak{v}}$, define
\[
\begin{aligned}
\DiamondOp_\Delta l &:= \{\,\DiamondOp_\Delta p: p\in l\,\},\\
\BoxOp_\Delta l &:= \{\,\BoxOp_\Delta p: p\in l\,\}.
\end{aligned}
\]
Then $\Truth{\DiamondOp_\Delta l}=\bigcap_{p\in l}\DiamondOp_\Delta p$ and similarly for $\BoxOp$.
\end{definition}

\Intuition To lift a statement, lift each ingredient and require all lifted ingredients to hold. With $\Diamond_\Delta$ this means every ingredient shows up somewhere in the window.

\begin{remark}[Temporal semantics as an abstraction layer]
The window environment $\Phi^\Delta$ and the lifted vocabulary $\{\DiamondOp_\Delta p: p\in\mathfrak{v}\}$ define a new abstraction layer in the native Stack-Theory sense, bridging ``over time'' constraints and pointwise semantics. The compilation of history-dependent (pathwise) abstractors into pointwise abstractors on augmented state is detailed in the Supplementary Material.
\end{remark}

\subsection{The Algebraic Core}
\label{sec:tg}

\begin{theorem}[Universal lift commutes with conjunction]
\label{thm:box}
For any statement $l\in\Lang{\mathfrak{v}}$ and any $\Delta\in\Nat$,
$\BoxOp_\Delta \Truth{l} = \Truth{\BoxOp_\Delta l}$.
\end{theorem}

\noindent\textit{Proof.} See Supplementary Material. $\Box_\Delta$ is ``for all positions''; since $\forall$ distributes over conjunction, $\Box_\Delta$ commutes with $\wedge$. \qed

\begin{theorem}[Existential lift does not commute with conjunction]
\label{thm:diamond}
For any statement $l\in\Lang{\mathfrak{v}}$ and any $\Delta\in\Nat$,
\[
\DiamondOp_\Delta \Truth{l}\;\subseteq\;\Truth{\DiamondOp_\Delta l}.
\]
Moreover, for $\Delta\ge 1$ there exist environments and statements for which the inclusion is strict.
\end{theorem}

\begin{proof}
If $\sigma\in\DiamondOp_\Delta\Truth{l}$ then for some $k$ I have $\sigma(k)\in\Truth{l}=\bigcap_{p\in l}p$.
Hence $\sigma\in\DiamondOp_\Delta p$ for every $p\in l$ and therefore $\sigma\in\bigcap_{p\in l}\DiamondOp_\Delta p=\Truth{\DiamondOp_\Delta l}$.

For strictness when $\Delta\ge 1$, let $\Phi=\{a,b,c\}$ and $\mathfrak{v}=\{p,q\}$ with $p=\{a,c\}$ and $q=\{b,c\}$.
Then $l=\{p,q\}\in\Lang{\mathfrak{v}}$ because $p\cap q=\{c\}\neq\emptyset$.
Take $\Delta=1$ and window $\sigma=(a,b)$.
Then $\sigma\in\DiamondOp_1 p$ and $\sigma\in\DiamondOp_1 q$, so $\sigma\in\Truth{\DiamondOp_1 l}$.
But $\sigma\notin\DiamondOp_1\Truth{l}=\DiamondOp_1(p\cap q)$ because neither $a$ nor $b$ lies in $p\cap q=\{c\}$.
\end{proof}

\ProofIntuition $\Diamond_\Delta$ is ``there exists a position''. Since $\exists$ does not distribute over conjunction, ingredients may occur at different positions without co-occurring.

\begin{remark}[Degenerate equality cases]
\label{rem:diamond_degenerate}
If $\Delta=0$ or $\card{l}\le 1$, then $\DiamondOp_\Delta \Truth{l} = \Truth{\DiamondOp_\Delta l}$.
Strict separation requires $\Delta\ge 1$ and at least two ingredients. A sufficient condition for commutation along a trajectory (within-window persistence) is given in the Supplementary Material.
\end{remark}

\begin{remark}[Reading the theorem back into window semantics]
In the sliding-window case, Definition~\ref{def:window_sat} corresponds to
\[
\begin{aligned}
\Occur_{\WinMap_{\Delta}}(l,\tau,t) &\iff \tau_{\Delta}(t)\in \Truth{\DiamondOp_\Delta l},\\
\CoInst_{\WinMap_{\Delta}}(l,\tau,t) &\iff \tau_{\Delta}(t)\in \DiamondOp_\Delta\Truth{l}.
\end{aligned}
\]
Theorem~\ref{thm:diamond} is exactly the claim that ingredient-wise satisfaction is strictly weaker than co-instantiation in general.
\end{remark}

\section{Chord and Arpeggio Postulates}
\label{sec:postulates}

The above algebra does not decide which window semantics corresponds to phenomenality.
I now state Chord and Arpeggio in terms of occurrence and co-instantiation, with a higher-layer moment statement and its base-layer grounding.

\begin{assumption}[Subjective moments live at some layer]
Fix a stack state with vocabularies $\mathfrak{v}^0,\dots,\mathfrak{v}^m$.
Assume a candidate subjective moment is represented by a conjunctive statement $l^m\in \Lang{\mathfrak{v}^m}$.
\end{assumption}

\begin{definition}[Phenomenal realisation predicate]
Let $\mathrm{PhenReal}(l^m,\tau,t)$ mean ``moment statement $l^m$ is phenomenally realised at (higher-layer) time $t$ along objective trajectory $\tau$.''
I treat $\mathrm{PhenReal}$ as primitive, constrained by Chord/Arpeggio.
\end{definition}

\Intuition $\mathrm{PhenReal}$ names the claim that moment $l^m$ is experienced at subjective time $t$ along run $\tau$. Here $l^m$ can include first, second, and third order selves and causal identities for objects and properties \citep{bennett2025thesis}.

\begin{definition}[Base grounding of a moment statement]
Let $g^0:=\Ground{0}{m}(l^m)$ be the base-layer grounding of $l^m$ (Definition~\ref{def:comp_ground}).
By Theorem~\ref{thm:comp_ground}, $\Truth{g^0}=\Truth{l^m}$.
I call $g^0$ the \emph{grounded conjunction} of the moment statement.
\end{definition}

\begin{definition}[Chord (objective co-instantiation required)]
\label{def:tgstrong}
Fix a windowing map for the phenomenal layer, typically the sliding window $\WinMap_{\Delta}$ of horizon $\Delta$ (Section~\ref{sec:window}).
Chord postulates that for all $l^m,\tau,t$,
\begin{align*}
\mathrm{PhenReal}(l^m,\tau,t) 
&\Rightarrow \CoInst_{\WinMap_{\Delta}}(g^0,\tau,t) \\
&\text{(equivalently } \tau_{\Delta}(t)\in \DiamondOp_\Delta\Truth{g^0}\text{).}
\end{align*}
If the moment is experienced, then within the corresponding subjective window there exists an objective time-slice at which the entire grounded conjunction holds.
\end{definition}

\Intuition Chord says: if the moment is experienced, then within the window there is at least one objective time step where all grounded ingredients are true together.

\begin{definition}[Arpeggio (objective smearing permitted)]
\label{def:tgweak}
Fix the same windowing map $\WinMap_{\Delta}$ as in Definition~\ref{def:tgstrong}.
Arpeggio postulates the following.
\begin{enumerate}[leftmargin=*, itemsep=0.25em]
\item \textbf{Ingredient-wise necessity.} For all $l^m,\tau,t$, \begin{align*}
&\mathrm{PhenReal}(l^m,\tau,t)\Rightarrow \Occur_{\WinMap_{\Delta}}(g^0,\tau,t) \\
&\text{(equivalently} \tau_{\Delta}(t)\in \Truth{\DiamondOp_\Delta g^0}\text{).}
\end{align*}
\item \textbf{Smearing permitted.} There exist $l^m,\tau,t$ such that
\[
\mathrm{PhenReal}(l^m,\tau,t)\wedge \Occur_{\WinMap_{\Delta}}(g^0,\tau,t)\wedge \neg \CoInst_{\WinMap_{\Delta}}(g^0,\tau,t).
\]
\end{enumerate}
This allows a moment to be experienced without any instant realising the full grounded conjunction.
\end{definition}

\Intuition Arpeggio says: if the moment is experienced, each grounded ingredient must occur somewhere in the window, but there need not be a single time step where they are all true together.

\section{Architectural Consequences}
\label{sec:arch}

A polycomputer simultaneously executes multiple functions using the same parts or physical substrate \citep{bongard2023}. A liquid brain such as an ant colony computes through movement rather than persistent structure \citep{sole2019}. In contrast, a ``solid brain'' has persistent structure supporting, e.g., a bio-electric abstraction layer with synchronous broadcast \citep{bennett2025thesis}.
I now translate ``single-thread sequential emulation'' versus ``synchronous polycomputational realisation'' into a minimal architectural invariant bounding \emph{simultaneous causal contribution}.
The Temporal Gap asks whether that bound is consciousness-relevant (and whether conscious unity requires causal exchange between co-instantiated ingredients, explored in \cite{bennett2026c}).

\subsection{Concurrency Capacity as an Invariant}

\begin{definition}[Contributor model]
Fix $n\ge 2$ potential contributors and write $[n]:=\{1,\dots,n\}$.
Define a base environment
$\Phi := X \times \Pow([n])$,
where $x\in X$ is a ``data'' component and $A\subseteq [n]$ is the set of currently \emph{active} contributors.
For each $i\in[n]$ define $p_i := \{(x,A)\in\Phi : i\in A\}$.
Let $l_n:=\{p_1,\dots,p_n\}$ and note $\Truth{l_n}=\{(x,[n]):x\in X\}$.
\end{definition}

\Intuition The contributor model records which contributors are active at each time step. A conjunction mentioning many contributors can only be true when all are active simultaneously.

\begin{definition}[Concurrency capacity]
An architecture has \emph{concurrency capacity} $c\in\{1,\dots,n\}$ if its admissible trajectories satisfy $\card{A_t}\le c$ for all objective times $t$ (writing $\tau(t)=(x_t,A_t)$).
The case $c=1$ formalises a sequential single-thread model; $c\ge n$ formalises synchronous polycomputing at the relevant scale.
\end{definition}

\Intuition Capacity $c$ is the maximum number of contributors active at one time step. If the grounded content needs more than $c$ active together, co-instantiation cannot hold.

\begin{theorem}[Synchronous threshold for co-instantiation]
\label{thm:capacity}
Fix $n\ge 2$ and content statement $l_n$.
If an architecture has concurrency capacity $c<n$, then along every admissible trajectory $\tau$ and for every horizon $\Delta$,
$\tau_{\Delta}(t)\notin \DiamondOp_\Delta\Truth{l_n}$ for all $t$.
However, if every contributor is activated at least once within some horizon-$\Delta$ window, then
$\tau_{\Delta}(t)\in \Truth{\DiamondOp_\Delta l_n}$.
Thus the architecture admits ingredient-wise satisfaction but forbids objective co-instantiation.
\end{theorem}

\begin{proof}
If $c<n$, then $\card{A_t}\le c<n$ at every $t$, so $A_t\neq[n]$ and $\tau(t)\notin\Truth{l_n}$ for all $t$.
No window contains an element of $\Truth{l_n}$, so $\DiamondOp_\Delta\Truth{l_n}$ is never satisfied.
For the second claim, if each $i$ appears in $A_{t+k_i}$ for some $k_i\le\Delta$, then $\sigma=\tau_{\Delta}(t)$ satisfies $\sigma\in\DiamondOp_\Delta p_i$ for all $i$, hence $\sigma\in\Truth{\DiamondOp_\Delta l_n}$.
\end{proof}

\ProofIntuition If at most $c$ contributors can be active, an $n$-way conjunction with $n>c$ is never true at an instant, though each ingredient can appear somewhere in a window.

\begin{corollary}[Single-thread CPU vs synchronous polycomputer]
\label{cor:cpu_poly}
Let a moment statement $l^m$ have base grounding $g^0$ whose ingredients include a conjunctive substatement equivalent to $l_n$ for some $n\ge 2$.
Fix the phenomenal windowing map $\WinMap_{\Delta}$ used in Chord/Arpeggio.
\begin{enumerate}[leftmargin=*, itemsep=0.35em]
\item Under Chord, any sequential-update architecture with capacity $c<n$ cannot phenomenally realise $l^m$ (Theorem~\ref{thm:capacity}).
\item Under Arpeggio, the same sequential architecture remains a candidate, because $\Truth{\DiamondOp_\Delta g^0}$ may hold without $\DiamondOp_\Delta\Truth{g^0}$.
\item Any synchronous architecture with capacity $c\ge n$ can satisfy $\DiamondOp_\Delta\Truth{g^0}$ by activating the required contributors simultaneously, hence can host the moment under both postulates.
\end{enumerate}
\end{corollary}

\begin{remark}[Connection to valence-first Stack Theory]
My valence-first application of Stack Theory treats certain phenomenal configurations as synchronised, causally efficacious conjunctions over many ingredients \citep{bennett2025thesis}.
The capacity model isolates one minimal implementation parameter---simultaneous contribution---controlling whether such conjunctions can be instantiated at any objective instant versus only ``smeared'' across a window.
\end{remark}

\section{Evidence for Chord}
\label{sec:neuro}

Chord and Arpeggio predict different empirical signatures \emph{within} a short subjective window. Either consciousness requires window-local coordinated co-engagement, or it can tolerate mere ingredient-wise activation distributed across the window.
Several classic paradigms are more naturally described by the coordination view.
In masking, conscious perception correlates with transient \emph{long-range} gamma phase synchrony across cortical areas, even when local activity is similar across seen versus unseen stimuli \citep{melloni2007synchronization}.
In non-REM sleep, TMS-EEG responses become strong yet \emph{local}, failing to propagate, consistent with breakdown of effective connectivity when consciousness fades \citep{massimini2005breakdown}.
Perturbational measures such as PCI track level of consciousness across wake/sleep/anesthesia and disorders of consciousness, consistent with a requirement for integrated, differentiated dynamics \citep{casali2013pci}.
These findings do not \textit{prove} Chord, but they make the existential permissiveness of Arpeggio less compelling. Conscious level covaries with the presence of a temporally coordinated, system-level regime.
I therefore treat Chord as the default working hypothesis; this step remains defeasible.

\section{Attribution and Implications}
\label{sec:agenda}

The Temporal Gap is the question of whether consciousness is affected by the \emph{gap between ingredient coverage and co-instantiation} for a grounded conjunction.

Fix a grounded conjunction $g^0$ and an objective trajectory $\tau$.
Define the minimal horizon needed for ingredient coverage versus co-instantiation:
\begin{align*}
w_{\mathrm{ing}}(g^0,\tau) &:= \inf\{\Delta\in\Nat:\exists t\ \tau_{\Delta}(t)\in\Truth{\DiamondOp_\Delta g^0}\},\\
w_{\mathrm{co}}(g^0,\tau) &:= \inf\{\Delta\in\Nat:\exists t\ \tau_{\Delta}(t)\in\DiamondOp_\Delta\Truth{g^0}\}.
\end{align*}
The strict Temporal Gap pattern is $w_{\mathrm{ing}}<w_{\mathrm{co}}$ (or $w_{\mathrm{co}}=\infty$).
These are \emph{instrumentable} when base-layer ingredients are defined in terms of architectural state variables.
This kind of architectural metric aligns with the view that behaviour alone may be insufficient for consciousness assessment and that concrete \emph{indicator properties} should be checked against candidate theories \citep{butlin2023ai}.

A concrete test follows: construct paired systems that are behaviourally matched at a reporting layer but differ in base-layer concurrency capacity (one synchronous with $c\ge n$, one sequential with small $c$). Select candidate contents whose groundings include large conjunctions, and measure $w_{\mathrm{ing}}$ and $w_{\mathrm{co}}$.
Chord predicts that phenomenality tracks bounded $w_{\mathrm{co}}$; Arpeggio predicts it may track $w_{\mathrm{ing}}$ alone.

If Chord is even \emph{plausible}, behavioural equivalence between a synchronous system and a sequential emulation does not settle phenomenal equivalence.
This creates an ethically salient zone of reasonable disagreement about moral status, particularly when a system could have valenced experience.
A precautionary stance toward sentience under uncertainty has been defended in nearby contexts \citep{birch2024edge}.
One concrete recommendation is to treat large Temporal-Gap regimes ($w_{\mathrm{co}}\gg w_{\mathrm{ing}}$ or $w_{\mathrm{co}}=\infty$) as a \emph{high-uncertainty} region for attribution, motivating conservative deployment policies until the Temporal Gap is better resolved.

\section{Conclusion}

I provided a Stack-Theory-compatible Stack-Time Semantics addendum and used it to make the Temporal Gap into a measurable algebraic and architectural question.
Compositional grounding (Theorem~\ref{thm:comp_ground}) ensures any higher-layer phenomenal statement has a base-layer grounded conjunction with identical truth conditions.
Induced layer time partitions objective time into maximal constant macro-blocks and can be arbitrarily sparse (Proposition~\ref{prop:sparse_ticks}), so subjective moments need not correspond to single objective instants.
The Temporal Gap becomes a question of the non-commutation pattern
\[
\DiamondOp_\Delta\Truth{l}\subseteq\Truth{\DiamondOp_\Delta l},
\]
with strict inclusion in general for $\Delta\ge 1$ and multi-ingredient statements (Theorem~\ref{thm:diamond}). This yields a clean separation between ingredient-wise window occurrence and objective co-instantiation.

Chord and Arpeggio are competing postulates about which notion constrains phenomenality.
Section~\ref{sec:neuro} notes that conscious level covaries with temporally coordinated dynamics, tentatively favouring Chord.

A simple architectural invariant (concurrency capacity) can force the strict Temporal Gap pattern for large conjunctions (Theorem~\ref{thm:capacity}).
Under Chord, software consciousness on a \emph{strictly sequential} substrate is impossible for any content whose grounding requires two or more simultaneous contributors.
Real machines are not perfectly sequential at the physical level, so the practical question becomes which physical concurrency scale is relevant for grounding. Consciousness according to Stack Theory requires a ``tapestry of valence'' supported by a solid-brained polycomputing architecture in which each co-instantiated ingredient simultaneously acts upon the others, all linked and synchronised (a condition formally included in Chord in \cite{bennett2026c}). Extreme concurrency capacity may be needed, and still concurrency is only one aspect of this. The central conclusion remains that consciousness attribution is hardware-sensitive, and surface-level behavioural equivalence alone cannot wash out the Temporal Gap.

\printbibliography

\end{document}